\documentclass[letterpaper, 10 pt, conference]{format/ieeeconf}  

\IEEEoverridecommandlockouts    
\usepackage{glossaries}
\newacronym{mav}{MAV}{Micro Aerial Vehicles}
\newacronym{uav}{UAV}{Unmanned Aerial Vehicle}
\newacronym{ovc}{OVC}{Open Vision Computer}
\newacronym{lidar}{LiDAR}{Light Detection and Ranging}
\newacronym{vio}{VIO}{visual-inertial odometry}
\newacronym{gpgpu}{GPGPU}{General-Purpose Graphics Processing Unit}
\newacronym{ugv}{UGV}{Unmanned Ground Vehicle}
\newacronym{uwb}{UWB}{Ultra Wideband}
\newacronym{svm}{SVM}{Support Vector Machine}
\newacronym{fcn}{FCN}{Fully Convolutional Network}
\newacronym{cnn}{CNN}{Convolutional Neural Network}
\newacronym{loam}{LOAM}{LiDAR Odometry and Mapping}
\newacronym{sloam}{SLOAM}{Semantic LiDAR Odometry and Mapping}
\newacronym{slam}{SLAM}{Simultaneous Localization and Mapping}
\newacronym{iot4ag}{IoT4Ag}{NSF Engineering Research Center for the Internet of Things for Precision Agriculture}
\newacronym{grasp-lab}{GRASP Lab}{the General Robotics, Automation, Sensing and Perception Laboratory}
\newacronym{jps}{JPS}{Jump Point Search}
\newacronym{ukf}{UKF}{Unscented Kalman Filter}
\newacronym{sam}{SAM}{Smoothing and Mapping}
\newacronym{icp}{ICP}{Iterative Closest Point}
\newacronym{imu}{IMU}{Inertial Measurement Unit}
\newacronym{tsdf}{TSDF}{Truncated Signed Distance Field}
\newacronym{esdf}{ESDF}{Euclidean Signed Distance Field}
\newacronym{sdf}{SDF}{Signed Distance Field}
\newacronym{rrt}{RRT}{Rapidly Exploring Random Tree}
\newacronym{fpv}{FPV}{First-person View}
\newacronym{dnn}{DNN}{Deep Neural Network}
\newacronym{igpred}{IGPred}{Information Gain Prediction}
\newacronym{csqmi}{CSQMI}{Cauchy-Schwarz Quadratic Mutual Information}
\newacronym{nbv}{NBV}{Next Best View}
\newacronym{vae}{VAE}{Variational Autoencoder}
\newacronym{tsp}{TSP}{Traveling Salesman Problem}
\newacronym{bcsm}{BCSM}{Behavior Control State Machine}
\newacronym{pca}{PCA}{Principal Component Analysis}
\newacronym{aspp}{ASPP}{Atrous Spatial Pyramid Pooling}
\newacronym{swap}{SWaP}{Size Weight and Power}
\newacronym{soi}{SoI}{Semantic Object of Interest}
\newacronym{aoi}{AoI}{Area of Interest}
\newacronym{drl}{DRL}{Deep Reinforcement Learning}
\newacronym{dl}{DL}{Deep Learning}
\newacronym{fov}{FoV}{Field of View}
\newacronym{tops}{TOPS}{Tera Operations per Second}

\usepackage[table,usenames,dvipsnames]{xcolor}      

\usepackage{tabularx}
\usepackage{multirow, multicol}
\usepackage{array}
\usepackage{tabu}
\usepackage{makecell}
\usepackage{adjustbox}
\usepackage{tikz}
\usetikzlibrary{shapes.geometric,arrows,positioning,calc}
\usepackage{float}

\newcolumntype{P}[1]{>{\centering\arraybackslash}p{#1}}
\newcolumntype{M}[1]{>{\centering\arraybackslash}m{#1}}
\usepackage{booktabs}
\newcolumntype{N}{>{\centering\arraybackslash}m{.5in}}
\newcolumntype{G}{>{\centering\arraybackslash}m{2in}}

\let\labelindent\relax

\usepackage{extarrows}                              
\usepackage{enumitem}
\usepackage{wrapfig}

\usepackage{amsmath,amssymb,amsfonts,amsthm,dsfont, bm} 
\usepackage{mathtools}
\usepackage{algorithm,algorithmicx,listings}        
\usepackage[noend]{algpseudocode}			              
\makeatletter
\def\BState{\State\hskip-\ALG@thistlm}
\makeatother
\DeclarePairedDelimiter\abs{\lvert}{\rvert}%
\DeclarePairedDelimiter\norm{\lVert}{\rVert}%

\makeatletter
\let\oldabs\abs
\def\abs{\@ifstar{\oldabs}{\oldabs*}}
\let\oldnorm\norm
\def\norm{\@ifstar{\oldnorm}{\oldnorm*}}
\makeatother

\usepackage{graphicx}
\usepackage{subcaption}
\usepackage{textcomp}
\usepackage[font=footnotesize]{caption}

\usepackage{xspace}
\usepackage{placeins}

\makeatletter
\DeclareRobustCommand\onedot{\futurelet\@let@token\@onedot}
\def\@onedot{\ifx\@let@token.\else.\null\fi\xspace}

\makeatother

\usepackage[normalem]{ulem}
\usepackage[utf8]{inputenc}
\usepackage[english]{babel}

\DeclareMathAlphabet\mathbfcal{OMS}{cmsy}{b}{n}

\newtheorem*{assumption*}{Assumption}

\newtheorem*{problem*}{Problem}

\usepackage{graphicx}

\usepackage{lipsum}

\makeatletter
\let\NAT@parse\undefined
\makeatother
\usepackage[space, compress, sort, noadjust]{cite}
\usepackage[breaklinks=true, colorlinks, bookmarks=true, citecolor=Black, urlcolor=Violet,linkcolor=Black]{hyperref}

\usepackage{arydshln}

\makeatletter
\DeclareMathSizes{9.95}{9}{7}{5}
\DeclareMathSizes{10}{9}{7}{5}
\DeclareMathSizes{11}{10}{8}{6}
\makeatother

\begin{document}

\title{
\textbf{HALO: High-Altitude Language-Conditioned\\
Monocular Aerial Exploration and Navigation}
}

\author{Yuezhan Tao*, Dexter Ong*, Fernando Cladera, Jason Hughes, \\Camillo J. Taylor, Pratik Chaudhari and Vijay Kumar
\thanks{*\textit{Yuezhan Tao and Dexter Ong contributed equally to this work.} The authors thank Alex Zhou for providing support for hardware platform. 
This work was supported by TILOS under NSF Grant CCR-2112665, IoT4Ag ERC under NSF Grant EEC-1941529, NSF grant CMMI-2415249, NSF NRI/USDA award 2022-67021-36856, the ARL DCIST CRA W911NF-17-2-0181, DSO National Laboratories and NVIDIA. All authors are with GRASP Laboratory, University of Pennsylvania {\tt\footnotesize\{yztao, odexter, fclad, jasonah, cjtaylor, pratikac, kumar\}@seas.upenn.edu}.} %
}

\maketitle

\begin{abstract}
We demonstrate real-time high-altitude aerial metric-semantic mapping and exploration using a monocular camera paired with a global positioning system (GPS) and an inertial measurement unit (IMU).
Our system, named HALO, addresses two key challenges: (i) real-time dense 3D reconstruction using vision at large distances, and (ii) mapping and exploration of large-scale outdoor environments with accurate scene geometry and semantics.
We demonstrate that HALO can plan informative paths that exploit this information to complete missions with multiple tasks specified in natural language.
In simulation-based evaluation across large-scale environments
of size up to 78,000 sq. m., HALO consistently completes tasks with less exploration time and achieves up to 68\% higher competitive ratio in terms of the distance traveled compared to the state-of-the-art semantic exploration baseline.
We use real-world experiments on a custom quadrotor platform to demonstrate that (i) all modules can run onboard the robot, and that (ii) in diverse environments HALO can support effective autonomous execution of missions 
covering up to 24,600 sq. m. area at an altitude of 40 m. 
Experiment
videos and more details can be found on our project page:
\url{https://tyuezhan.github.io/halo/}.
\end{abstract}


\section{Introduction}
\label{sec:intro}

Aerial robots operating at high altitudes have a large effective field-of-view, this can be used very effectively for mapping and exploration.
However, high-altitude aerial operations present some unusual challenges in perception.
For example, consumer-grade LiDARs provide accurate depth but the point density at large distances is low.
LiDARs are also expensive, heavy and do not provide the same richness of information as cameras.
Vision-based systems are also more attractive because they are inexpensive and lightweight.
Vision is however difficult to utilize for high-altitude aerial autonomy.
Stereo systems are ineffective due to the limited baseline on aerial vehicles~\cite{warren2013high} while monocular systems have scale ambiguities.
Global positioning systems (GPS) provides accurate position information but not the orientation.
These challenges make it difficult to use existing hardware solutions in high-altitude aerial applications.

Traditional geometric mapping approaches for robot autonomy, while sufficient for basic navigation tasks, fall short of providing the rich contextual understanding required for complex mission planning and human-robot interaction in real-world scenarios.
Advances in vision-language models (VLMs) have shown promising results in robotics for providing semantically rich representations that facilitate behaviors driven by natural language.
However, extending these capabilities to high-altitude aerial platforms introduces additional complexity due to the computational constraints of onboard processing and the nature of the aerial imagery.

This paper addresses the above challenges for high-altitude aerial semantic exploration with tasks specified in natural language. The proposed system, which we call HALO, incorporates recent advances in feed-forward 3D reconstruction and dense vision-language representations to build open-set metric-semantic maps from aerial imagery. This is coupled with a hierarchical planning architecture that balances exploration and exploitation for efficient semantic navigation on a high-altitude aerial platform.

\begin{figure}
  \centering
  \includegraphics[width=\linewidth]{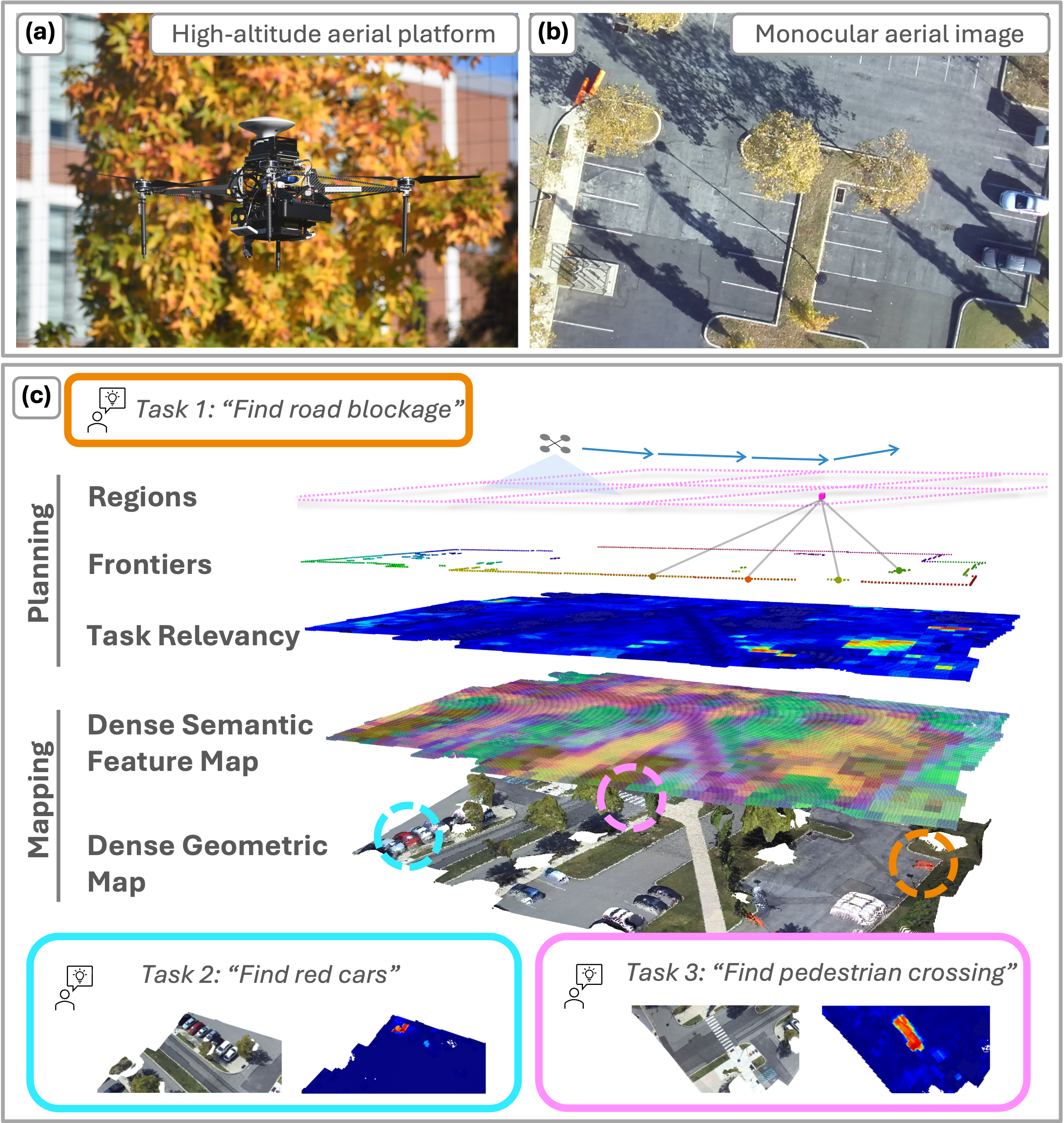}
  \caption{\textbf{Conceptual overview of HALO:} A high-altitude aerial platform (a) captures monocular aerial imagery (b) as it navigates the environment. HALO constructs a metric-semantic map in real-time (c).
  Upon this map, a hierarchical planner detects semantically relevant frontiers to make informative plans that balance exploration and exploitation to complete tasks specified in natural language.
  The semantic map contains dense language embeddings that provide the flexibility to adapt to new tasks with a wide range of semantics in real time.
  Three tasks are presented in the map above in orange, cyan, and magenta.}
  \label{fig:placeholder}
  \vspace{-0.4cm}
\end{figure}

This is the first high-altitude monocular robotic system that demonstrates dense language-embedded metric-semantic mapping with language-conditioned exploration in real time onboard the robot.%
\footnote{All code, scripts and system schematics will be made public imminently.}
The contributions of this paper are as follows.

\begin{enumerate}
\item \textbf{An approach for real-time monocular open-set metric-semantic mapping from high-altitude aerial imagery:}
We fuse predictions from a feed-forward 3D reconstruction model with classical graph-based mapping and localization techniques for dense geometric reconstruction.
We detect geometric frontiers from the 3D reconstruction, enabling occlusion-aware 3D reasoning.
We construct a semantic map that stores dense language features which can provide context beyond object-level scene understanding even in unstructured environments, while maintaining the flexibility to support a diverse range of tasks.

\item \textbf{A hierarchical planning framework for efficient semantic navigation:}
We propose a hierarchical planning framework for task-relevant information gathering.
The global planner leverages geometric and semantic information from the map to identify regions of interest, maintaining a balance between semantic exploration and exploitation.
The local planner interprets this global plan by generating either an optimal information-gathering exploration path or a shortest information-retrieval exploitation path.

\item \textbf{An autonomous aerial robotic system for language-conditioned exploration and navigation at high altitudes:}
We incorporated the aforementioned mapping and planning modules into an autonomous aerial robotic platform.
We validated our system through extensive simulation experiments with multiple baselines as well as real-world experiments running in real time onboard the robot across a diverse range of tasks.
\end{enumerate}

\section{Related Work}
\label{sec:related work}

\subsection{High-altitude Aerial Mapping}

The most common use of aerial images for real-time applications is to build orthomosaics~\cite{peterson2018online, he2025multisensorfusionapproachrapid,hinzmann2017mapping} by stitching 2D aerial images.
Other methods build sparse semantic maps from aerial imagery~\cite{cladera2025air} or embed closed-set semantics in the orthomaps~\cite{miller2022stronger}.
However, these methods have a limited potential for scene understanding because they do not represent geometric information or are limited to a small set of semantic classes.
Some approaches incorporate depth from a stereo camera~\cite{wagner2022online} or from monocular visual SLAM~\cite{kern2020openrealm,miller2024air} to create elevation maps, though these approaches are either sparse or limited in the quality of stereo depth.
Several works~\cite{2013diasmultistereo,2022shaomars} aim to overcome these limitations with multi-robot collaborative stereo.
Structure-from-Motion (SfM) can be used for orthomapping to obtain dense point clouds~\cite{geographies4010005}. However, this is computationally expensive and has to be done offline.
We aim to achieve vision-based high-altitude aerial dense mapping with dense language embeddings using a monocular camera in real time onboard a UAV.

\subsection{Geometric and Semantic Exploration}
Autonomous exploration has been extensively studied over the past few decades. 
Frameworks that rely on geometric cues from the environment have been developed to guide exploration. 
Yamauchi~\cite{yamauchi1997frontier} introduced the use of frontiers, which represent the boundary between known and unknown space, to direct exploration into unexplored regions. 
More recently, this concept has been extended with additional heuristics to evaluate and prioritize frontier visitation~\cite{zhou2021fuel, RapidExploration, ral24_active_3d_slam}.
Information-driven methods, which are typically based on the entropy of the geometric map, offer another heuristic for guiding exploration~\cite{charrow2015CSQMI, KelseyIG, LukasIG}. 

With advances in semantic segmentation and VLMs, semantic information has also been leveraged to guide exploration and target search~\cite{georgakis2021learning, yuezhantao2023seer, asgharivaskasi2023semantic, yokoyama2024vlfm,ong2025atlas, alama2025rayfronts}. 
In~\cite{georgakis2021learning}, the authors presented an approach to predict semantic maps beyond the field of view and used semantic uncertainty to guide exploration. 
SEER~\cite{yuezhantao2023seer} employed semantic cues in indoor environments to inform exploration strategies. 
The work in~\cite{asgharivaskasi2023semantic} presented an information-driven exploration framework based on a semantic-augmented octomap.
VLFM~\cite{yokoyama2024vlfm} generates a value map with a VLM to rank frontiers based on the occupancy map. ATLAS~\cite{ong2025atlas} and Rayfronts~\cite{alama2025rayfronts} further incorporate language features into Gaussian Splatting and voxel maps, enabling semantic navigation and exploration.
However, most existing methods that leverage open-vocabulary semantics for exploration are designed for ground robots, where computational resources and latency constraints are less restrictive. Moreover, these approaches often balance memory requirements against map quality by compressing semantic features~\cite{ong2025atlas} and embedding them into dense geometric maps. In contrast, our framework embeds dense semantic features to enable large-scale, language-conditioned exploration with high-altitude aerial robots.

\subsection{3D Reconstruction with Learned Geometric Priors}
Feed-forward 3D reconstruction models have brought about a significant step forward in unified, end-to-end deep learning frameworks for 3D reconstruction.
DUST3R~\cite{wang2024dust3r} and its variants operate on image pairs with joint encoders and separate decoders to predict pointmaps.
VGGT~\cite{wang2025vggt} improves multi-view correspondence and geometric consistency across input images with a camera head for camera parameters and a dense prediction transformer for dense depth and tracking.
Several approaches~\cite{murai2024_mast3rslam,cut3r,maggio2025vggt,deng2025vggtlong,streamVGGT} have leveraged these models for incremental mapping and SLAM.
These methods open up exciting possibilities in situations where stereo depth estimation fails or where accurate pose information is difficult to acquire.
In this work, we explore the use of such models for real-time, incremental dense 3D reconstruction from high-altitude monocular aerial imagery.

\section{Problem Specification}
\label{sec:formulation}
Given an unknown environment and mission specifications that may include multiple tasks revealed sequentially rather than defined \textit{a priori}, our objective is to enable real-time efficient autonomous exploration and task execution with a high-altitude aerial robot. Efficiency is quantified by the competitive ratio between the executed path length and the shortest possible path required to complete the task. 
This problem requires the robot to (i) construct and maintain an accurate map from high altitude images and (ii) make online decisions to adaptively acquire task-relevant information through either exploration of new areas or exploitation of the existing map.

\section{Methods}
\label{sec:method}

\begin{figure*}[t]
\centering
    \vspace{0.1cm}
\includegraphics[width=0.98\linewidth]{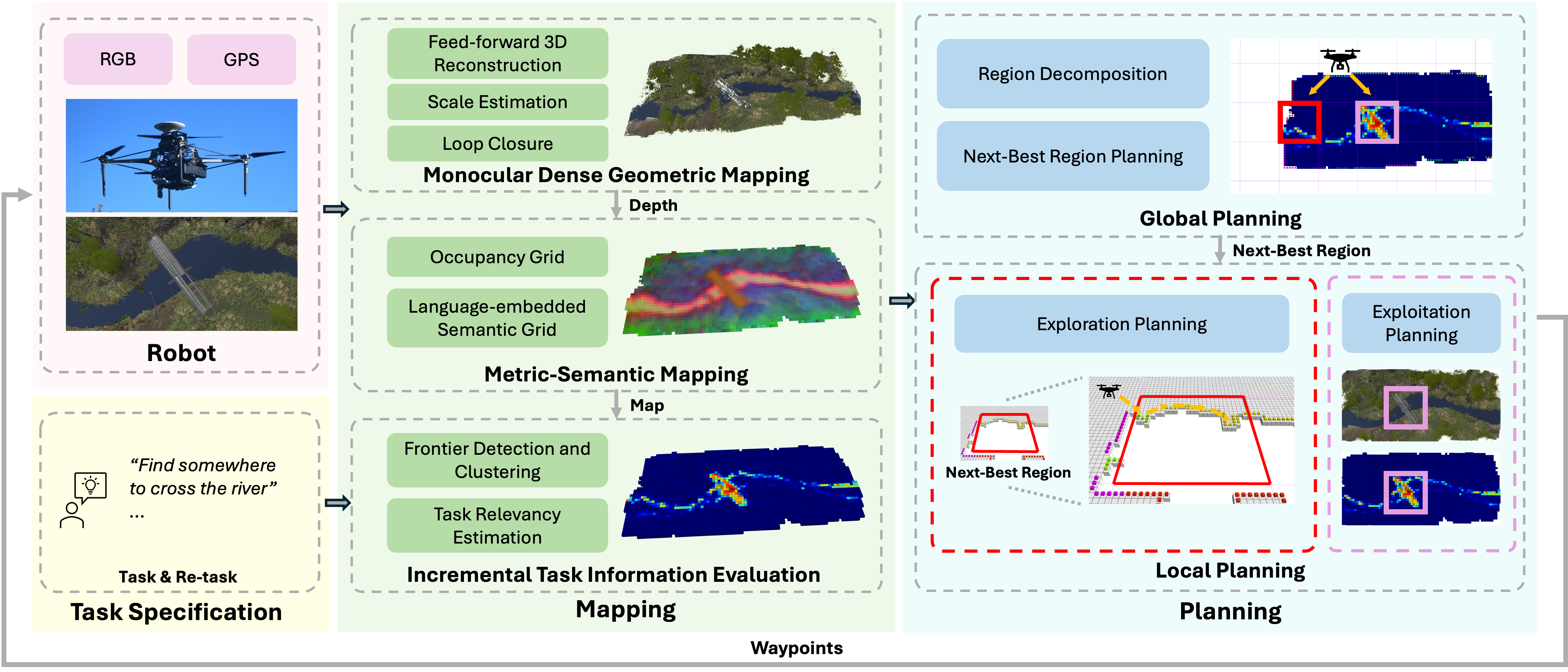}
\caption{\textbf{System overview of HALO:}
The robot takes in monocular RGB images and GPS measurements.
The monocular dense geometric mapping module estimates depth camera poses from a sequence of images.
The metric-semantic mapping module maintains an occupancy grid and a dense semantic feature grid.
Frontiers and task relevancy are extracted from the metric-semantic map incrementally. 
Leveraging both geometric and semantic information, the hierarchical planner balances exploration and exploitation to plan paths for efficient task completion.
}
\label{fig:system_diagram}
\vspace{-0.4cm}
\end{figure*}

This section focuses on the two key parts of HALO to enable this: (i) a mapping module that maintains dense geometric information along with language embeddings, and (ii) a hierarchical planning module that consists of a high-level planner to select the next-best region along with a local planner to generate paths for information gathering or retrieval. 
An overview of the system is illustrated in Fig.~\ref{fig:system_diagram}.

\subsection{High-altitude Aerial Metric-semantic Mapping}
\label{subsec:mapping}
We design an incremental mapping module based on VGGT for estimating dense depth and poses from high-altitude aerial imagery. This information will then be used for semantic mapping and exploration.

\begin{figure}[t]
  \centering
  \resizebox{\linewidth}{!}{
    \begin{tikzpicture}[
      pose/.style={circle,draw=gray!80,fill=gray!20,thick,minimum size=15mm},
      prior/.style={rectangle,draw=magenta!80,fill=magenta!20,thick,minimum size=8mm},
      icp_factor/.style={rectangle,draw=cyan!80,fill=cyan!20,thick,minimum height=6mm,minimum width=12mm,font=\small},
      vggt_factor/.style={rectangle,draw=blue!80,fill=blue!20,thick,minimum height=6mm,minimum width=12mm,font=\small},
      loop_factor/.style={rectangle,draw=orange!80,fill=orange!20,thick,minimum height=6mm,minimum width=12mm,font=\small},
      node distance=2cm and 3cm
      ]

    \node[pose] (x1) {$x_1$};
    \node[pose, right=of x1] (x2) {$x_2$};
    \node[pose, right=of x2] (x3) {$x_3$};
    \node[pose, right=of x3] (xn) {$x_n$};

    \node[prior, above=1cm of x1] (p1) {G};
    \draw[-] (p1) -- (x1);
    \node[prior, above=1cm of x2] (p2) {G};
    \draw[-] (p2) -- (x2);
    \node[prior, above=1cm of x3] (p3) {G};
    \draw[-] (p3) -- (x3);
    \node[prior, above=1cm of xn] (p4) {G};
    \draw[-] (p4) -- (xn);

    \node[icp_factor, above=7mm of $(x1)!0.5!(x2)$, xshift=0mm] (b12_1) {ICP};
    \node[vggt_factor] (b12_2) at ($(x1)!0.5!(x2)$) {F3DR};
    \draw[-] (x1) -- (b12_1);
    \draw[-] (x2) -- (b12_1);
    \draw[-] (x1) -- (b12_2);
    \draw[-] (x2) -- (b12_2);

    \node[icp_factor, above=7mm of $(x2)!0.5!(x3)$, xshift=0mm] (b23_1) {ICP};
    \node[vggt_factor] (b23_2) at ($(x2)!0.5!(x3)$) {F3DR};
    \draw[-] (x2) -- (b23_1);
    \draw[-] (x3) -- (b23_1);
    \draw[-] (x2) -- (b23_2);
    \draw[-] (x3) -- (b23_2);

    \node at ($(x3)!0.5!(xn)$) {...};
    \node[loop_factor, below=14mm of $(x2)!0.5!(xn)$, xshift=0mm] (b2n_1) {F3DR};
    \draw[-] (x2) -- (b2n_1);
    \draw[-] (xn) -- (b2n_1);
    \end{tikzpicture}
  }
  \caption{\textbf{Monocular dense aerial mapping using feed-forward 3D reconstruction models (F3DR).} GPS readings are used to provide position priors G (magenta). The pose graph also maintains ICP (cyan) and F3DR (blue) relative pose estimates as constraints between submaps. Loop closures are implemented with relative pose estimates from F3DR (orange).}
  \label{fig:vggt_factor_graph}
  \vspace{-0.4cm}
\end{figure}
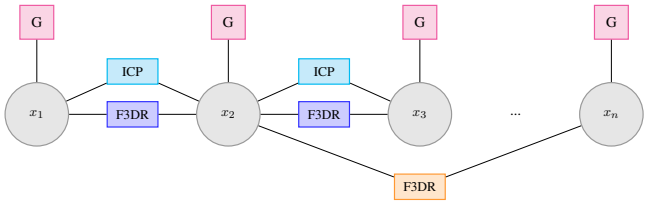

\medskip \noindent \textbf{High-altitude monocular geometric mapping}
Fig.~\ref{fig:vggt_factor_graph} provides an overview of the VGGT-based mapping module to estimate camera intrinsics, dense depth and camera poses.
Similar to VGGT-SLAM, we process batches of images to build submaps and align these submaps using a pose graph. There are two constraints to connect poses across submaps.
The first is on the scaled relative pose of images (obtained from VGGT) that were used to build the same submap.
We get RGB-colored point clouds from VGGT, which allow us to calculate a relative transformation between two adjacent submaps with colored Iterative Closest Point (ICP)~\cite{park2017colored}.
To facilitate real-time operation, we process images in small batches.
While VGGT-SLAM implements only one frame of overlap between submaps, we use a larger number of overlapping frames to improve ICP alignment.
We also found empirically that running ICP over smaller batches results in better alignment, making this strategy well-suited for incremental mapping.

GPS measurements do not provide orientation and can be noisy, but the positional information can complement approaches like VGGT which predict orientation accurately but suffer from scale ambiguity in translation.
We use GPS measurements as position priors in the pose graph.
We also estimate a scale factor on the VGGT position estimates using the GPS priors.
VGGT depth and pose estimates are scaled before being added to the pose graph.
Our approach thus recovers scale using VGGT predictions and reduces long-term positional drift in the submap pose graph.

Similar to VGGT-SLAM, we implement loop closure detection using SALAD descriptors~\cite{izquierdo2024optimal}.
We only select submaps within a certain threshold from the current location as candidates for loop closure detection to reduce the overhead of checking all submaps at every iteration.
The best loop closure candidate frame is added to the current batch.
The pose estimate from VGGT is added to the pose graph as a relative pose constraint between the current submap and the loop closure candidate submap.
We do not use ICP as an additional constraint for the loop closure because there can be situations when the corresponding point clouds might not overlap even when loop closure is detected.

\medskip \noindent \textbf{Language-embedded metric-semantic mapping}
We next use depth images from the preceding dense mapping module, together with the corresponding RGB images and estimated poses to construct 2D metric-semantic grid maps. Our semantic map consists of dense language features embedded in a 2D grid, it can provide open-set semantic guidance for aerial exploration.
The 2D semantic grid, as opposed to a full 3D map which we could have also built in principle, allows us to keep the memory requirements manageable.

\uline{Occupancy grid:}
At high altitude, the robot is always operating in free space, and thus a representation that distinguishes only between known and unknown regions is sufficient for our problem, i.e., we do not maintain occupancy probabilities.
Initially, all grid cells in the 2D occupancy map are set to unknown.
The map is updated by back-projecting depth measurements from the camera frame and discretized into grid cells.
Any cell observed through the projected 2D measurements is marked as known.

\uline{Semantic feature grid:}
We extract per-pixel language features from each color image.
We use a modified RADIO~\cite{heinrich2025radiov25improvedbaselinesagglomerative} encoder with a SigLIP~\cite{zhai2023sigmoid} adapter to obtain dense language-aligned features as in~\cite{hajimiri2025naclip,alama2025rayfronts}.
We back-project the feature map to a point cloud, where each point contains an associated language feature vector.
These embeddings are aggregated into the corresponding 2D grid cells.
When we have multiple observations of the same grid cell, we use exponential moving average to update its features.

\medskip \noindent \textbf{Incremental task information evaluation}
We develop a task information evaluation module that incrementally computes the utility of various regions of the map to guide task-driven exploration or exploitation.

\uline{Frontier detection and clustering}
We compute geometric frontiers on the occupancy grid map.
Consider a bounding box that encapsulates the region of the map that changed after the latest map update.
Within this bounding box, we first clear all existing frontiers, and then detect new frontier clusters using a region-growing algorithm.
Clusters with a radius larger than $\text{ftr}_{\text{min}}$ are retained whereas those exceeding $\text{ftr}_{\text{max}}$ are iteratively subdivided along their first principal axis until then they are smaller.
The bounding box for computing geometric frontiers is reinitialized after each map update.

\uline{Estimating task relevance}
We maintain a grid of task relevance scores that are computed from the semantic feature grid.
For a given task, we first obtain text embeddings of the task description using the SigLIP text encoder.
We then compute the cosine similarity between the task text embeddings and the semantic feature embeddings to obtain a relevancy score for each grid cell in the semantic feature grid.
Similar to the occupancy grid map updates, for each map update we only need to compute the relevancy of cells within the updated bounding box.
The next section describes how this task-relevancy map is used for scoring regions and frontiers in the global and local planner, respectively.

\subsection{Hierarchical Planning}
\label{subsec:planning}
The planner in HALO decides whether to expand the map to acquire new information or to revisit previously explored regions to exploit existing information. To this end, we develop a two-level planning module that
consists of a global planner and a local planner. 
This hierarchical planning module operates on the frontiers of the geometric map and the task-relevancy map to generate paths for efficient task-driven navigation.

\medskip \noindent \textbf{Global planning}
The high-level planner decomposes the task area into 2D regions of size $s_{\text{reg}} \times s_{\text{reg}}$.
Regions are assigned frontier clusters based on whether the geometric centroid of the cluster lie within that region.
Regions containing at least one frontier are labeled as ``exploration regions'' $R_e$.
The utility $u$ of an exploration region is computed by averaging the task relevancy (after thresholding the semantic relevancy map with $\epsilon_{e}$) of the grid cells in the region.

When a new task is received, the planner identifies the grid cell with the highest task relevancy in each fully-explored region and compares it to a threshold $\epsilon_r$.
Regions with task relevancy higher than $\epsilon_r$ are labeled as ``exploitation regions'' $R_r$.
A region that is marked for exploitation remains so until it is actually visited by the robot, or until a new task arises.

For each candidate region in the set $R_e \cup R_r$, we calculate the Euclidean distance between the robot pose and the center of the region.
The next-best region is the one that maximizes the cost-benefit ratio $\text{argmax}_i u_i/c_i$ for all $i \in R_e \cup R_r$.
The global planner selects the next-best region at a fixed rate.

\medskip \noindent \textbf{Local planning}
The local planner is invoked in two cases, either when a global plan is updated, or in a receding horizon fashion while following a global plan.
The local planner receives the next-best region and its label from the global planner, and computes a shortest path to the region based on the label.
The label determines the behavior of the robot: if labeled for exploration, the path is designed to gather new information, if labeled for exploitation, the path is designed for exploitation of previously gathered information.

\uline{Exploration-based path planning}
We set up an Asymmetric Traveling Salesman Problem (ATSP) to explore a region as follows.
We first filter the semantic relevancy map using a threshold $\epsilon_{\text{ftr}}$ to obtain utility scores.
The average utility of each frontier cluster is then computed, and clusters with mean utility below $\epsilon_{\text{ftr}}$ are pruned out.
This step ensures that the robot only visits frontier clusters that are likely to provide task-relevant information.
If no frontier clusters provide sufficient utility, all clusters within the region are retained.
This is because the task-relevant area within the region may be far from existing frontier clusters.

We then construct a directed planning graph with $N_{\text{ftr}} + 1$ nodes, where the $N_{\text{ftr}}$ nodes each represent a centroid $p$ of a candidate frontier cluster after the pruning step above, together with a node representing the robot location $x$.
The cost matrix $C$ is a $(N_{\text{ftr}}+1) \times (N_{\text{ftr}}+1)$ square matrix that represents the edges between these nodes.
Costs of edges between frontier clusters are the Euclidean distance between their corresponding centroids:
\(
    C(i, j) = C(j,i) = ||p_i -p_j||_2
\)
for $i,j \in \{1, 2, \ldots N_{\text{ftr}}\}$.
Similar to FUEL~\cite{zhou2021fuel}, we set the outbound edge from the robot node to any frontier clusters as the Euclidean distance between them, and the inbound edge to be zero cost.
This makes it possible to remove the edge that returns to the robot location in the solution without changing the total cost of the solution.
This approach computes an optimal path that visits all of the $N_{\text{ftr}}$ frontiers.

\uline{Exploitation-based path planning}
If the next-best region is an exploitation region, it indicates that there exists a previously visited grid cell therein that is strongly relevant to the current task. In such cases, the robot should simply visit that cell/region again.
In all our experiments, the size of the region are set according to the field-of-view and altitude of the robot.
For exploitation regions, the path directly connects the current location of the robot to the centroid of the region.

\begin{table}[!t]
\centering
    \vspace{0.1cm}
\caption{\textbf{Reconstruction error for 3D reconstructions from high-altitude flights in simulation built incrementally.} Failure of the method is indicated by a ``--''.}
\begin{adjustbox}{width=\linewidth}
\renewcommand{\arraystretch}{1.5}
\begin{tabular}{l rrr rrr}
    \toprule
    \multirow{4}{*}{\textbf{Method}}
    & \multicolumn{3}{c}{\textbf{PolyCity}}
    & \multicolumn{3}{c}{\textbf{Forest}} \\
    & \multicolumn{3}{c}{(1,604 m, 78,000 m$^2$)}
    & \multicolumn{3}{c}{(517 m, 62,500 m$^2$)} \\
    \cline{2-7}
    & Acc. & Compl. & Chamf. & Acc. & Compl. & Chamf. \\
    & {[m] $\downarrow$} & {[m] $\downarrow$} & {[m] $\downarrow$}
    & {[m] $\downarrow$} & {[m] $\downarrow$} & {[m] $\downarrow$} \\
    \midrule
    MASt3R-SLAM~\cite{murai2024_mast3rslam} & 53.22 & 20.11 & 55.69 & -- & -- & -- \\
    VGGT-SLAM-SIM(3)~\cite{maggio2025vggt} & \cellcolor{red!40}\textbf{8.29} & \cellcolor{orange!40}3.58 & \cellcolor{red!40}\textbf{9.09} & 8.37 & \cellcolor{orange!40}2.36 & \cellcolor{orange!40}7.10 \\
    VGGT-Long~\cite{deng2025vggtlong} & 33.41 & 17.73 & 48.10 & \cellcolor{red!40}\textbf{4.66} & 5.51 & 8.32 \\
    Ours (without GPS) & \cellcolor{orange!40}11.30 & \cellcolor{red!40}\textbf{2.62} & \cellcolor{orange!40}11.12 & \cellcolor{orange!40}7.95 & \cellcolor{red!40}\textbf{2.15} & \cellcolor{red!40}\textbf{7.01} \\
    \bottomrule
\end{tabular}
\end{adjustbox}
\label{tab:vggt_ablation}
\vspace{-0.4cm}
\end{table}

\begin{table*}
\centering
    \vspace{0.1cm}
\caption{\textbf{Quantitative results of simulation experiments.}
PolyCity missions: Easy -- find a blue car; Med. -- find somewhere to have a picnic; Hard -- [find a car accident, find a bench, find the football stadium].
Forest missions: Easy -- find a house along the trail; Med. -- find somewhere to cross the river; Hard -- [find a lake, find a bridge, find a house near the lake].
}
\begin{adjustbox}{width=0.9\linewidth}
\renewcommand{\arraystretch}{1.2}
\begin{tabular}{l rrrr rrrr rrrr}
\toprule
\multirow{3}{*}{\textbf{Method}} & \multicolumn{6}{c}{{\textbf{PolyCity}}} & \multicolumn{6}{c}{{\textbf{Forest}}}\\
& \multicolumn{3}{c}{\textbf{Time [s] $\downarrow$} } & \multicolumn{3}{c}{\textbf{Competitive Ratio $\uparrow$}} &
\multicolumn{3}{c}{\textbf{Time [s] $\downarrow$}} & \multicolumn{3}{c}{\textbf{Competitive Ratio $\uparrow$}} \\
& Easy & Med.  & Hard & Easy & Med.  & Hard & Easy & Med.  & Hard & Easy & Med.  & Hard \\
\midrule

Coverage & 420.3 & 403.8 & 895.2 & 0.180 & 0.154 & 0.080 & 237.6 & 370.3 & 1073.9 & 0.218 & 0.156 & 0.069 \\
Frontier~\cite{yamauchi1997frontier} & \cellcolor{orange!40}298.4 & 346.8 & 804.4 & 0.251 & 0.212 & 0.119 & 245.0 & 298.2 & 689.8 & 0.218 & 0.224 & 0.112 \\

FUEL~\cite{zhou2021fuel} & 312.7 & 896.6 & 1410.3 & \cellcolor{orange!40}0.345 & 0.092 & 0.082 & 504.1 & 580.1 & 835.5 & 0.088 & 0.117 & 0.112 \\

VLFM~\cite{yokoyama2024vlfm} & 334.1 & \cellcolor{orange!40}230.2 & \cellcolor{orange!40}457.9 & 0.214 & \cellcolor{orange!40}0.275 & \cellcolor{orange!40}0.173 & \cellcolor{red!40}\textbf{148.2} & \cellcolor{orange!40}203.5 & \cellcolor{orange!40}118.2 & \cellcolor{orange!40}0.375 & \cellcolor{orange!40}0.373 & \cellcolor{orange!40}{0.708}\\

Ours & \cellcolor{red!40}\textbf{239.5} & \cellcolor{red!40}\textbf{161.7} & \cellcolor{red!40}\textbf{401.6} & \cellcolor{red!40}\textbf{0.361} & \cellcolor{red!40}\textbf{0.343} & \cellcolor{red!40}\textbf{0.189} & \cellcolor{orange!40}154.2 & \cellcolor{red!40}\textbf{172.6} & \cellcolor{red!40}\textbf{99.8} & \cellcolor{red!40}\textbf{0.378} & \cellcolor{red!40}\textbf{0.427} & \cellcolor{red!40}\textbf{0.795} \\
\bottomrule
\end{tabular}
\end{adjustbox}
\vspace{-0.4cm}
\label{tab:sim_results}
\end{table*}


\section{Simulation Experiments}

We use a custom Unity-based simulator to independently evaluate the effectiveness of the monocular geometric mapping module and the planning in HALO.
The simulator can simulate a high-altitude quadrotor to provide aerial imagery along with depth and pose for obtaining groundtruth 3D reconstructions.
We used two environments for our experiments---PolyCity and Forest. PolyCity represents a typical urban environment with roads, vehicles, buildings, and other urban structures.
Forest is a largely unstructured environment with sparse road and river networks and several man-made structures like cabins and bridges.
We implemented a ROS 2 interface between the simulator and our autonomy framework.

\subsection{High-altitude Monocular Geometric Mapping}

We compared our approach to five other methods that use feed-forward 3D reconstruction methods for incremental 3D reconstruction from monocular images---MASt3R-SLAM~\cite{murai2024_mast3rslam}, CUT3R~\cite{cut3r}, StreamVGGT~\cite{streamVGGT}, VGGT-SLAM~\cite{maggio2025vggt} and VGGT-Long~\cite{deng2025vggtlong}.
We ran VGGT-SLAM with both the SL(4) and SIM(3) optimization approaches presented in their work.

\begin{figure}[!t]
\centering
\begin{adjustbox}{width=0.98\linewidth}
\setlength{\tabcolsep}{0em}
\begin{tabular}{ccccc}
    & {\scriptsize VGGT-SLAM-SIM(3)} & {\scriptsize VGGT-Long} & {\scriptsize Ours (w/o GPS)} & {\scriptsize GT} \\
    \raisebox{0.4\height}{\rotatebox{90}{\scriptsize PolyCity}} &
    \includegraphics[width=0.25\linewidth]{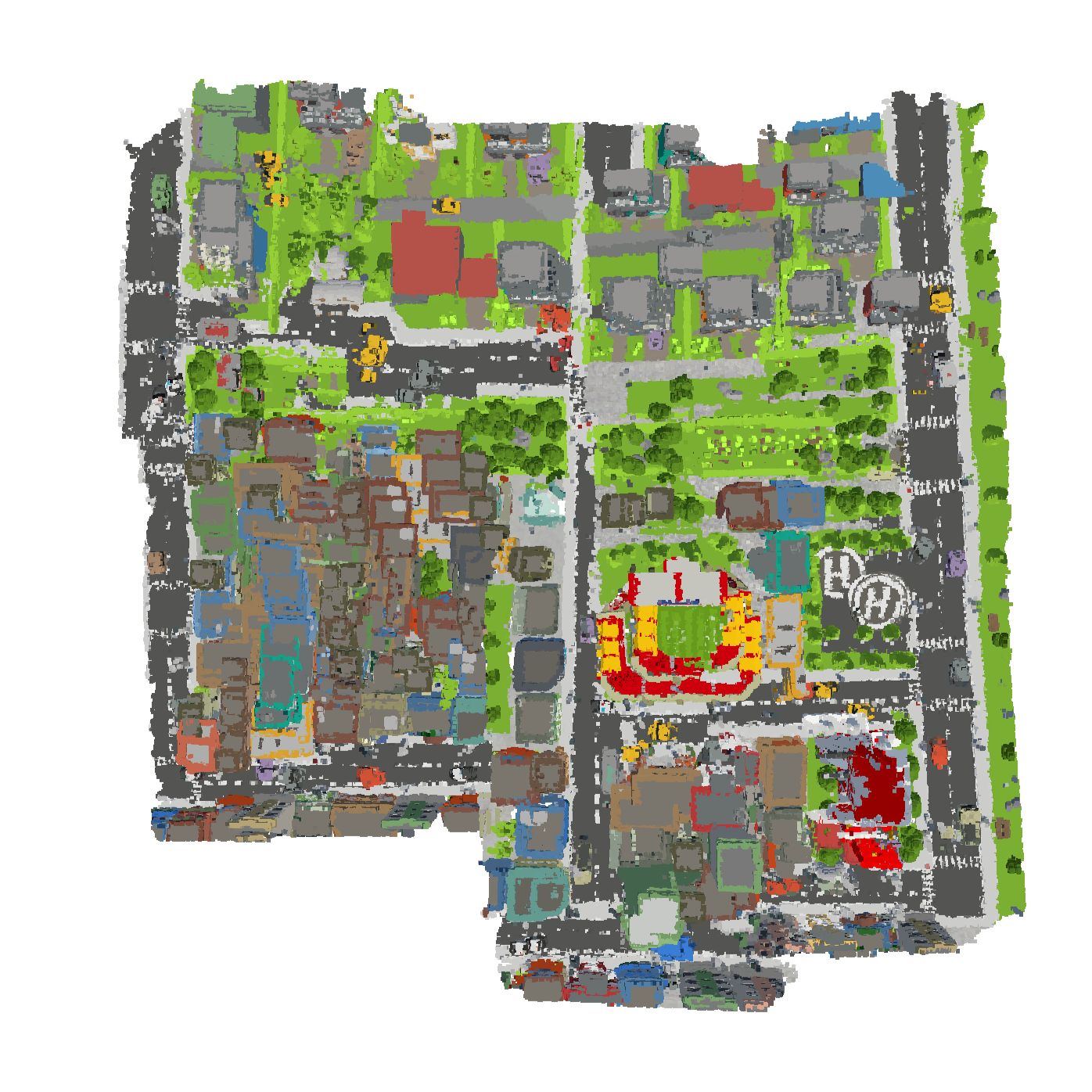} &
    \includegraphics[width=0.25\linewidth]{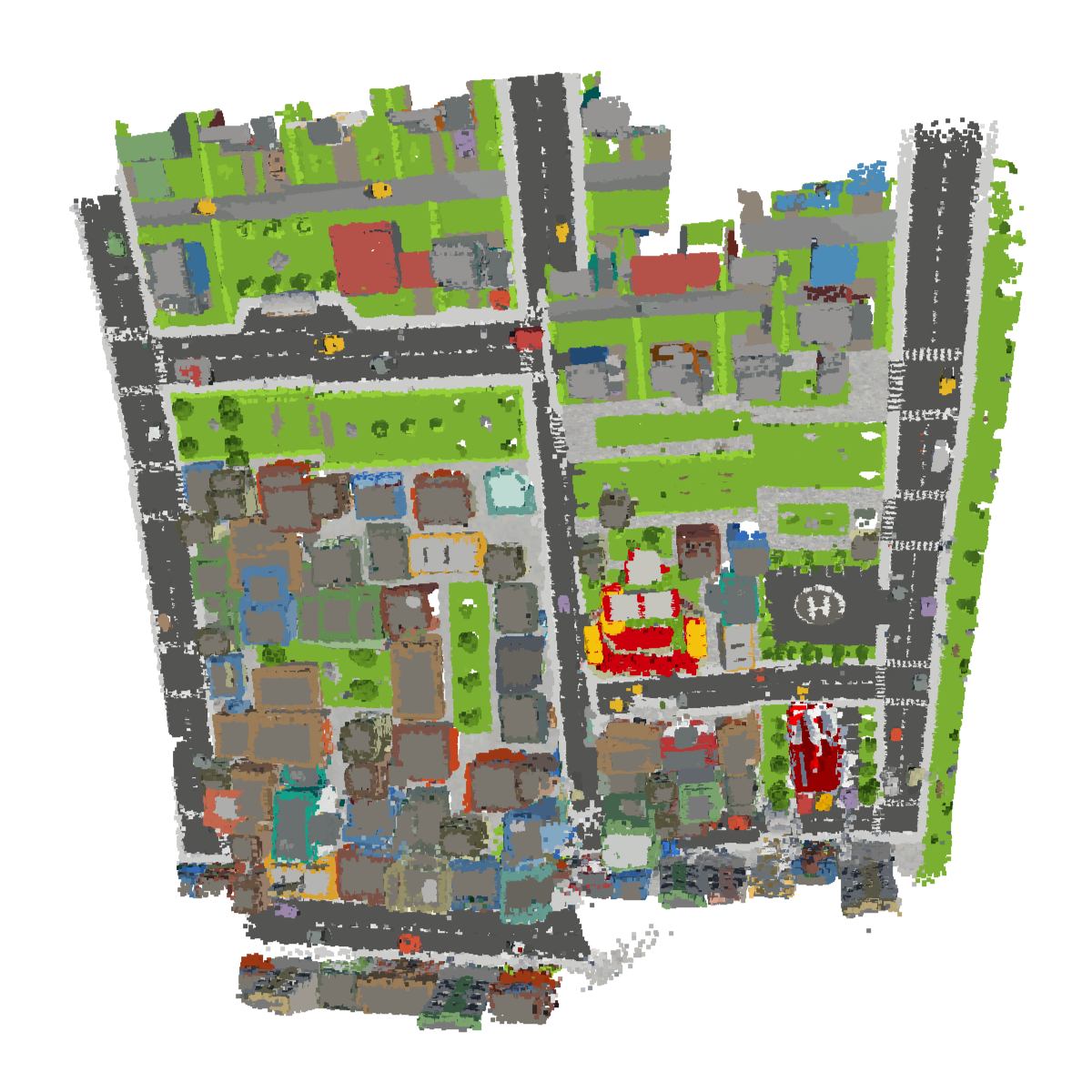} &
    \includegraphics[width=0.25\linewidth]{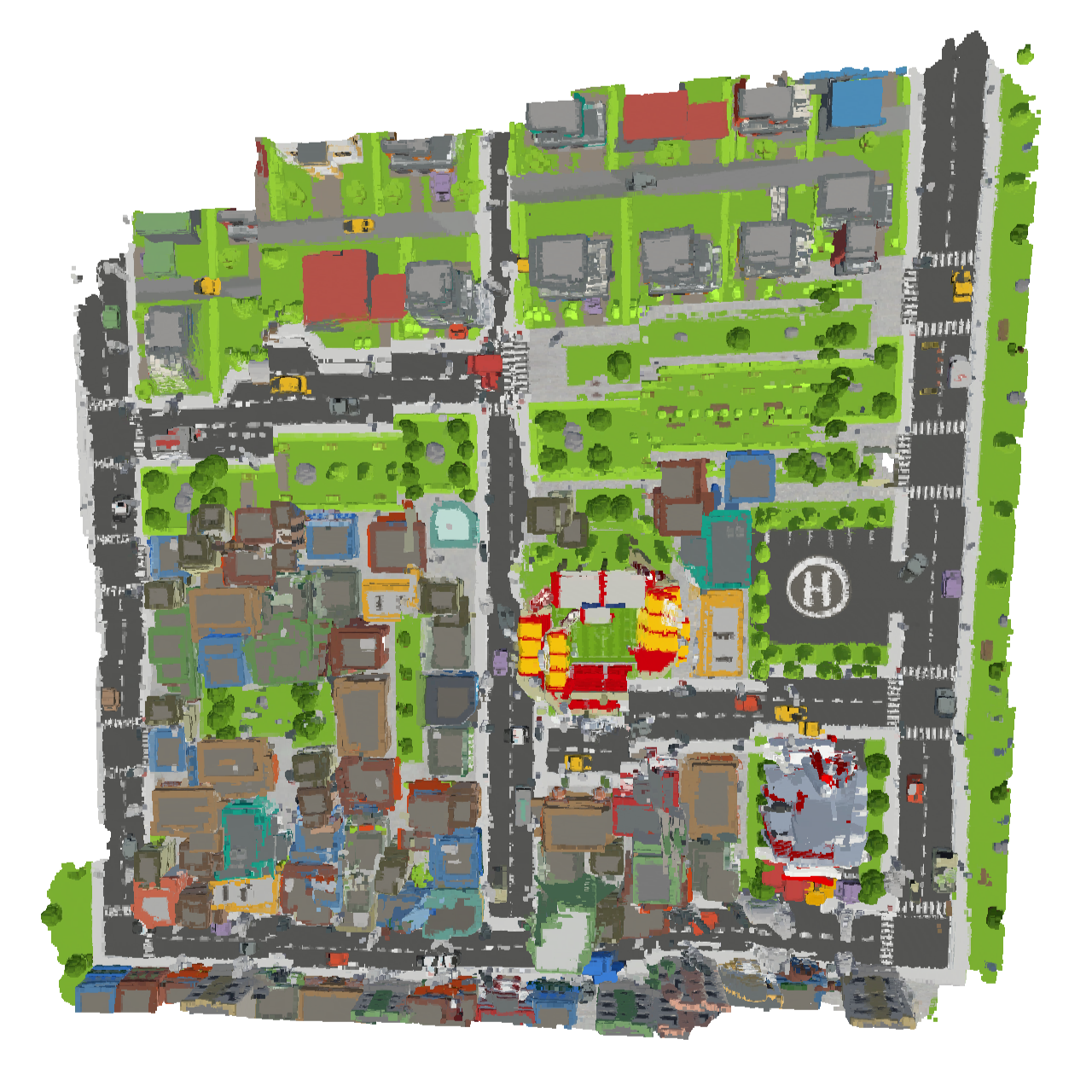} &
    \includegraphics[width=0.25\linewidth]{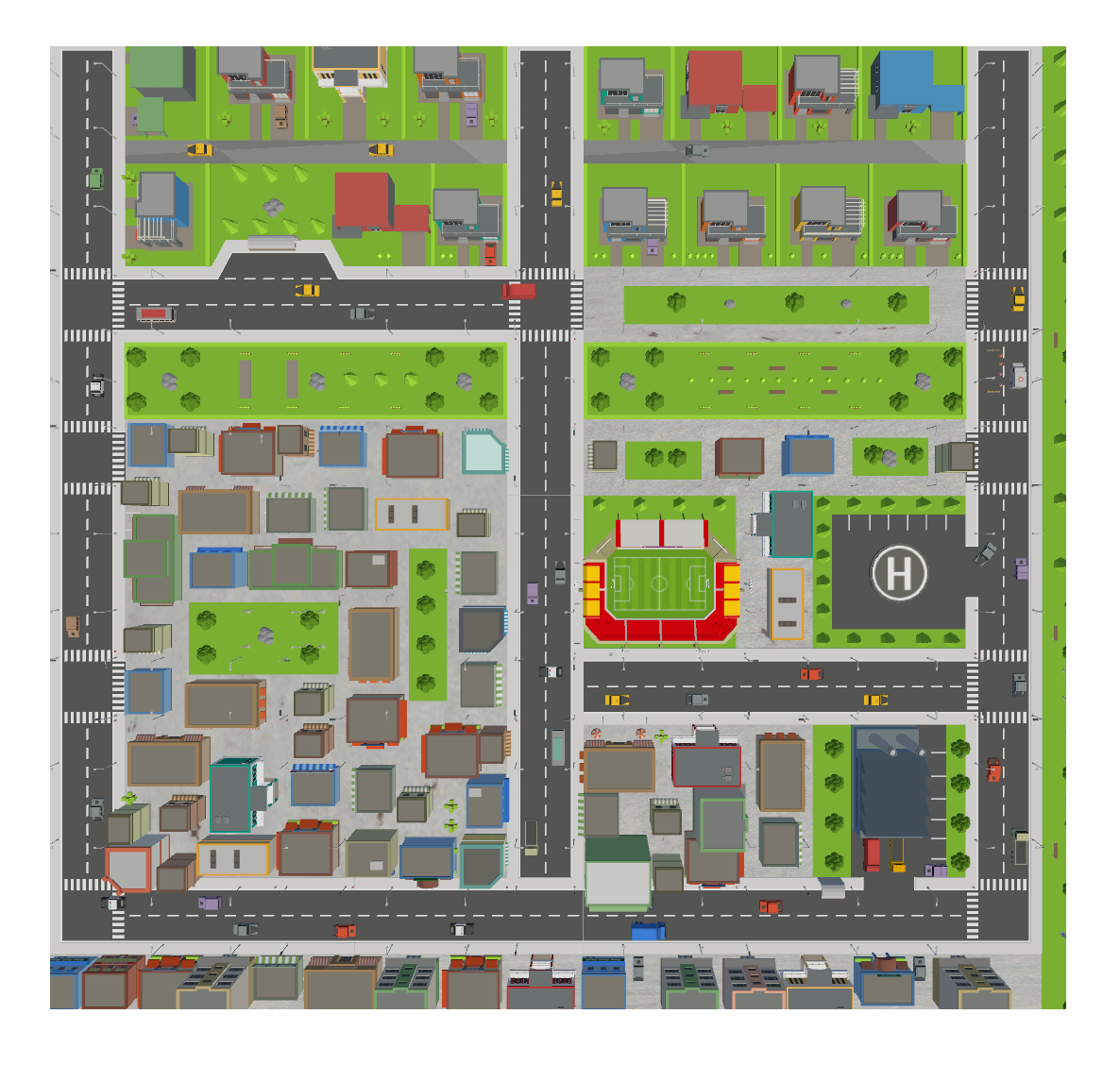} \\
    \raisebox{1.2\height}{\rotatebox{90}{\scriptsize Forest}} &
    \includegraphics[width=0.25\linewidth]{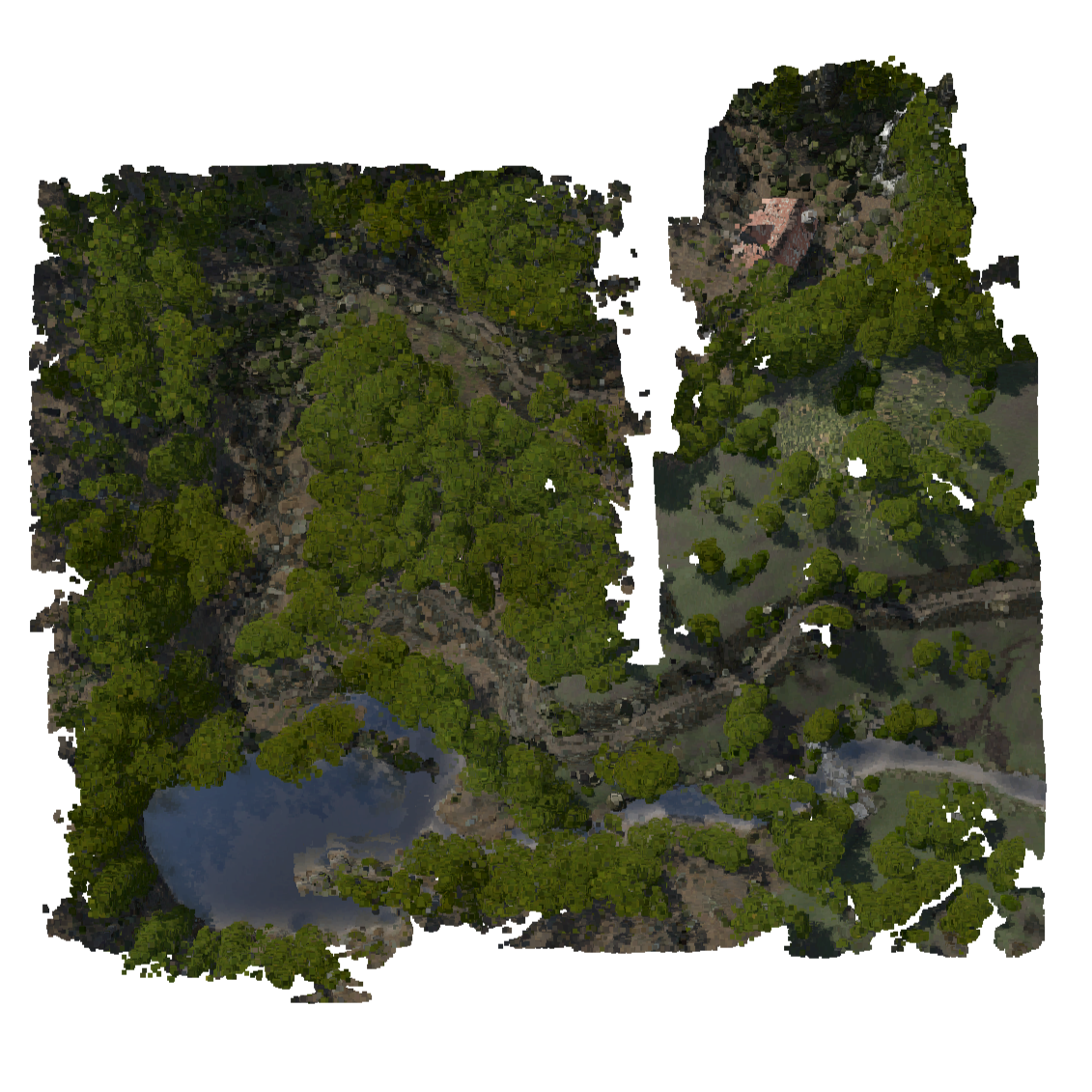} &
    \includegraphics[width=0.25\linewidth]{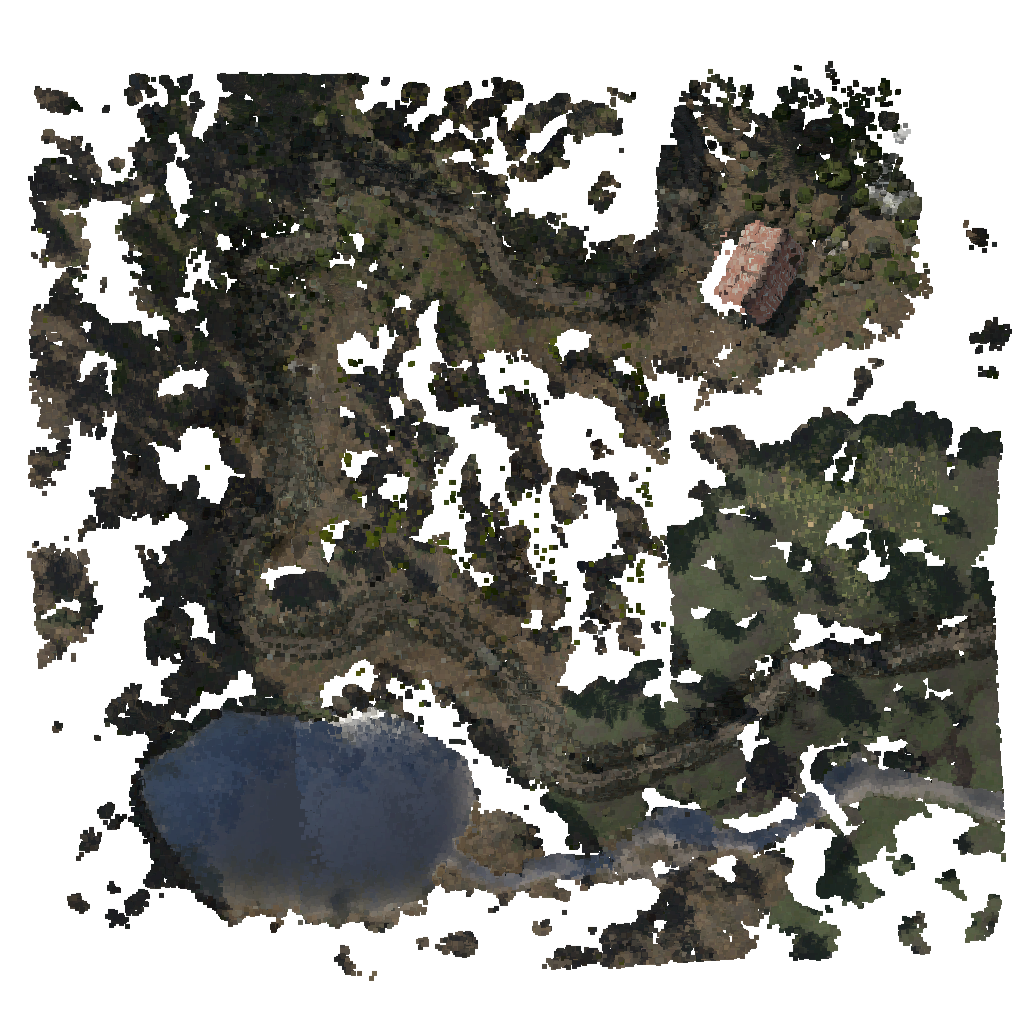} &
    \includegraphics[width=0.25\linewidth]{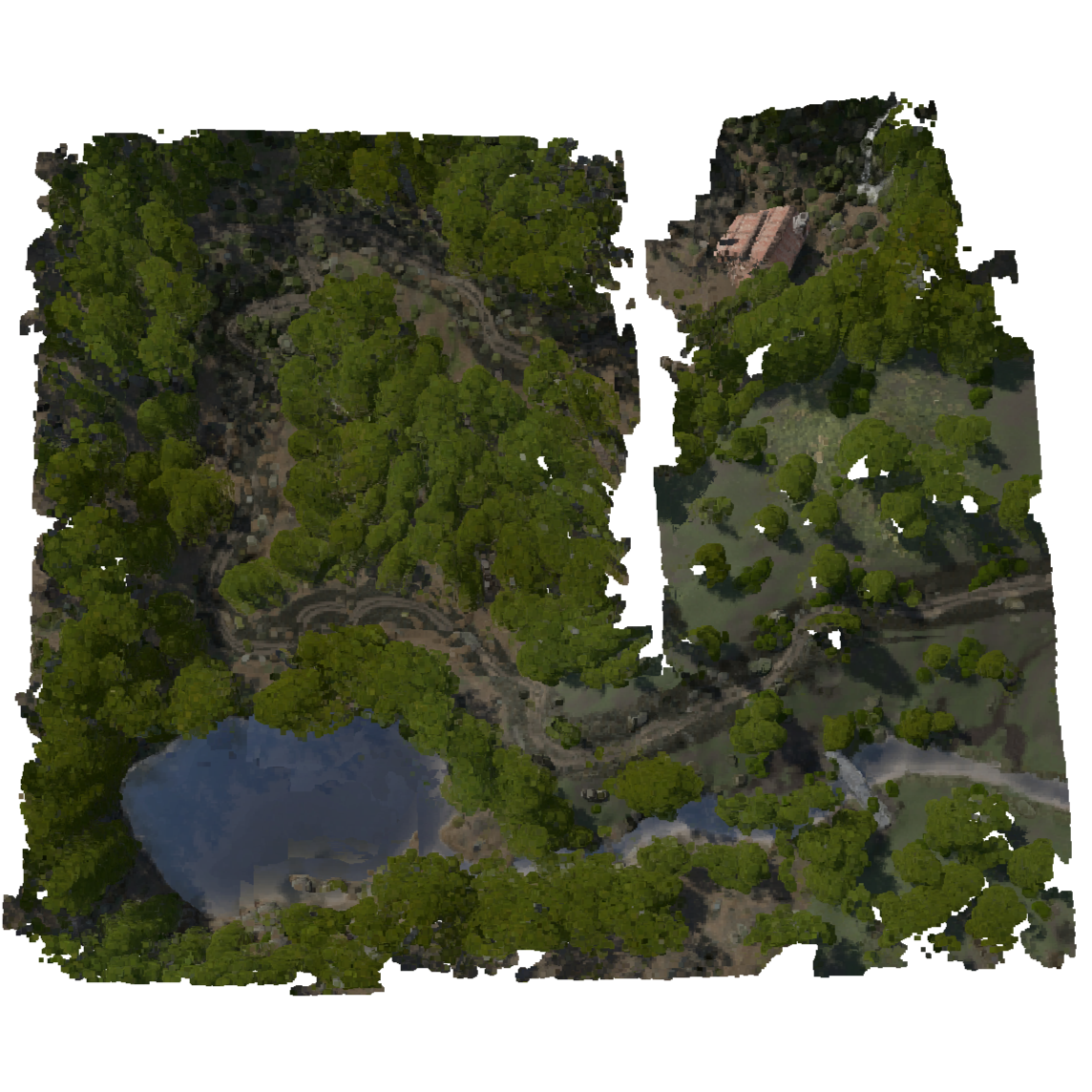} &
    \includegraphics[width=0.25\linewidth]{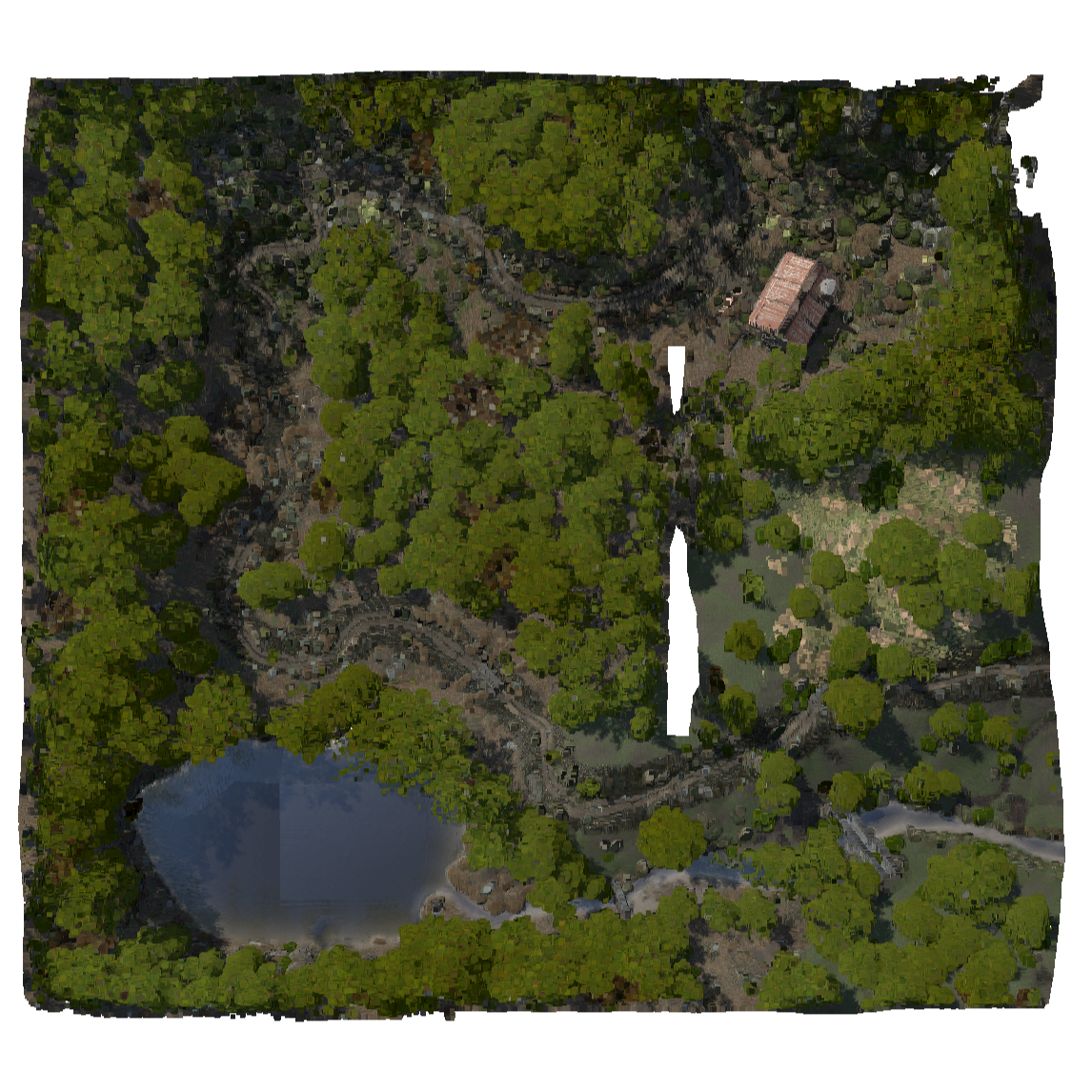} \\
\end{tabular}
\end{adjustbox}
\caption{\textbf{Visualization of 3D reconstructions of various methods on PolyCity and Forest in Unity.}}
\vspace{-0.4cm}
\label{fig:vggt_pointclouds}
\end{figure}

For each environment, we ran a boustrophedon coverage path and collected images every 2 m along the trajectory.
The trajectories generated are of lengths 1604 m and 517 m and cover areas of 78,000 m$^2$ and 62,500 m$^2$ respectively.
The ground truth point cloud was obtained from depth images and poses in the simulator.
All methods use the same set of images, with 5 images per submap.
For a fair comparison, we do not use GPS priors in our method and compute a scale between adjacent submaps as done in~\cite{maggio2025vggt}.
We scaled each reconstructed point cloud to the ground truth point cloud and compute three point cloud error metrics:
(i) accuracy, which is the root mean square error (RMSE) of the distance between each point in the reconstructed point cloud and the nearest ground truth point,
(ii) completion is the distance between each ground truth point and the nearest reconstructed point, and
(iii) Chamfer distance, which is the averaged two-sided distance.
Tab.~\ref{tab:vggt_ablation} presents the results while Fig.~\ref{fig:vggt_pointclouds} provides visualizations.

CUT3R, StreamVGGT and VGGT-SLAM-SL(4) failed completely on both scenes and are not presented in the table.
CUT3R and StreamVGGT do not generalize well to aerial images in our experiments.
MASt3R-SLAM provides reasonable reconstruction for PolyCity but it has significant drift, and it completely failed for Forest.
Compared to VGGT-SLAM-SIM(3), our method reconstructs the buildings in PolyCity and the lake in Forest more accurately, as shown in Fig.~\ref{fig:vggt_pointclouds}.
Though these metrics do not fully capture this, preserving such structure is essential for semantic navigation.
The global optimization of VGGT-Long leads to good structure, although there is significant rotation drift in PolyCity.
However, global optimization makes VGGT-Long unsuitable for real-time incremental mapping.
Without metric accurate scale, these approaches cannot be used for real-world navigation.

\subsection{Metric-semantic Exploration}

\medskip \noindent \textbf{Baselines}
To evaluate the performance of our semantic planning module, we conducted experiments that compare HALO to four other baseline methods. These include classical geometric exploration approaches such as Coverage, Frontier~\cite{yamauchi1997frontier}, and FUEL~\cite{zhou2021fuel}, as well as state-of-the-art semantic navigation methods like VLFM~\cite{yokoyama2024vlfm}.
The Coverage baseline directs the robot to the nearest corner of the environment and executes a boustrophedon path to survey the space, given the field-of-view and altitude of the robot.
The classical Frontier planner~\cite{yamauchi1997frontier} greedily directs the robot to the nearest frontier cluster.
For FUEL~\cite{zhou2021fuel}, we follow the original implementation, which plans a traversal path through all frontier clusters, but exclude costs related to heading.
For VLFM~\cite{yokoyama2024vlfm}, which assigns utility to frontiers and selects the best candidate based on a cost–benefit trade-off, we use utility scores from our relevancy map.

\medskip \noindent \textbf{Environments and Missions}
For both PolyCity and Forest, we designed three types of missions, shown in Tab.~\ref{tab:sim_results}, with varying levels of difficulty that is based on the number of tasks and the specificity of the task descriptions.
We evaluated all methods on two variations of the hard mission: one with three mutually relevant tasks, and another with three unrelated tasks.
The hard-relevant mission was tested in the Forest environment, and the hard-irrelevant mission in the PolyCity environment.
To ensure a thorough evaluation, we selected 2 distinct start locations in each environment and executed each method 5 times per mission at each start location---this leads to a total of 300 simulation runs.

\medskip \noindent \textbf{Analysis}
Tab.~\ref{tab:sim_results} presents the results from these simulation experiments.
Broadly, we observe that methods that use both geometric and semantic information consistently outperform those that rely only on the former.
In certain cases, geometric methods perform comparably, because incorporating semantic information can occasionally lead to unnecessary exploration.
This typically occurs when the goal is nearby but the semantic map does not provide obvious guidance.

The performance of semantics-driven methods remains relatively stable, even when the task is under-specified, this is seen by observing the competitive ratio (ratio of the optimal ground truth path length and the actual traversed distance) for easy and medium difficulty missions.
For the hard mission with unrelated tasks (PolyCity), HALO can execute a sequence of semantic-driven tasks even when previously explored areas might not have contained information that is useful for subsequent tasks.
In such cases, HALO achieves only moderate improvements, with 12\% and 50\% reductions in exploration time, and 9\% and 59\% improvement in competitive ratio, compared to VLFM and the best baseline geometric approach, respectively.
In contrast, for the hard mission with related tasks (Forest), our planner effectively exploits the semantic embeddings obtained during exploration for efficient information retrieval, resulting in a much higher competitive ratio.
As shown in Tab.~\ref{tab:sim_results}, HALO achieves over 15\% and 85\% reductions in exploration time, and 12\% and more than six times improvements in competitive ratio, compared to VLFM and the best geometric baseline, respectively.
Compared with the state-of-the-art semantic-driven approach~\cite{yokoyama2024vlfm}, HALO achieves improvements in competitive ratio ranging from 9.2\% to 68.7\% in the PolyCity environment and up to 15\% in the Forest environment.

\begin{figure*}
  \centering
      \vspace{0.1cm}
  \includegraphics[width=0.98\linewidth]{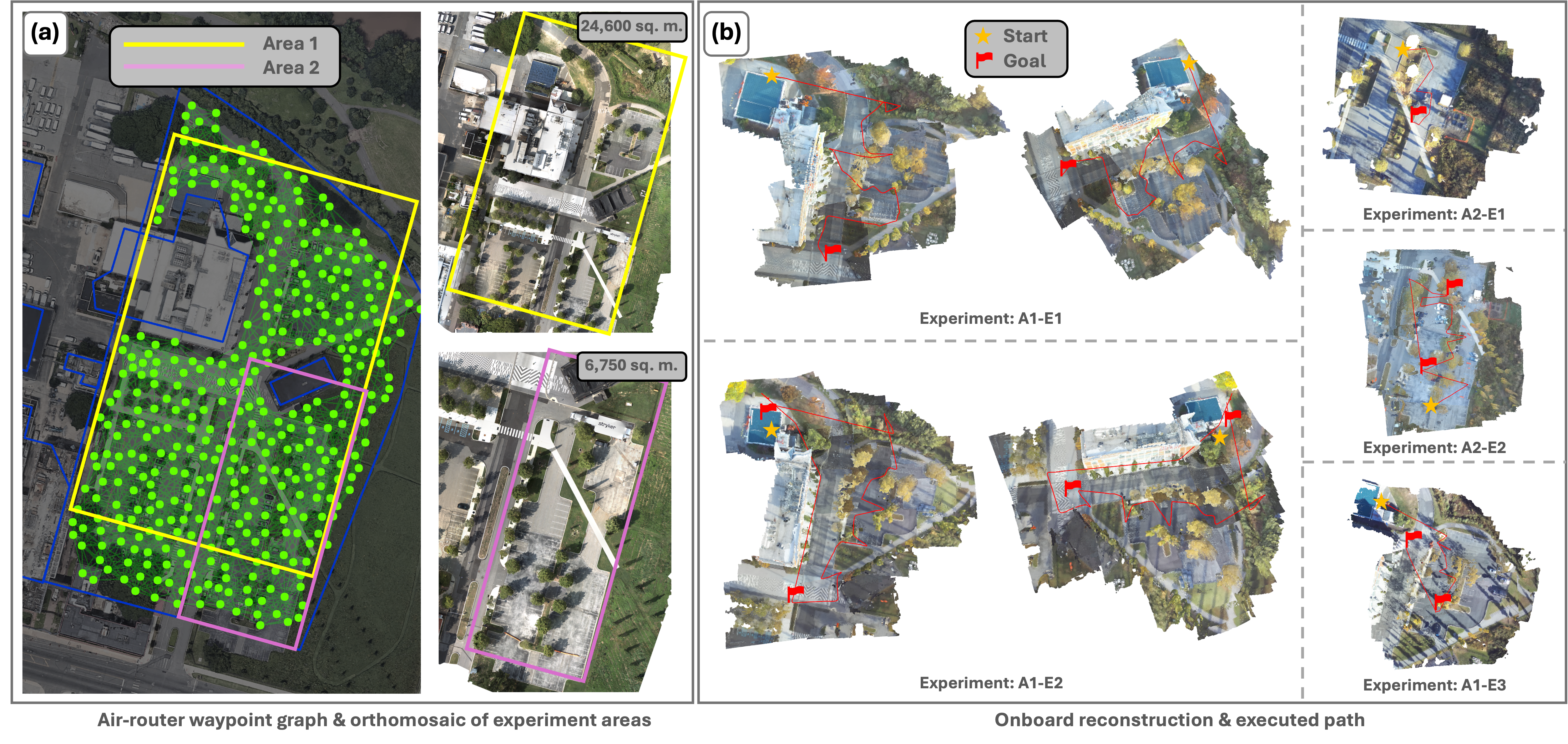}
  \caption{\textbf{Qualitative results from real-world experiments.} The safe and reachable waypoints and experiment areas are shown in (a). The onboard reconstructed maps are shown in (b), with executed paths in red.}
  \label{fig:experiment}
  \vspace{-0.5cm}
\end{figure*}

\begin{figure}
  \centering
  \includegraphics[width=\linewidth]{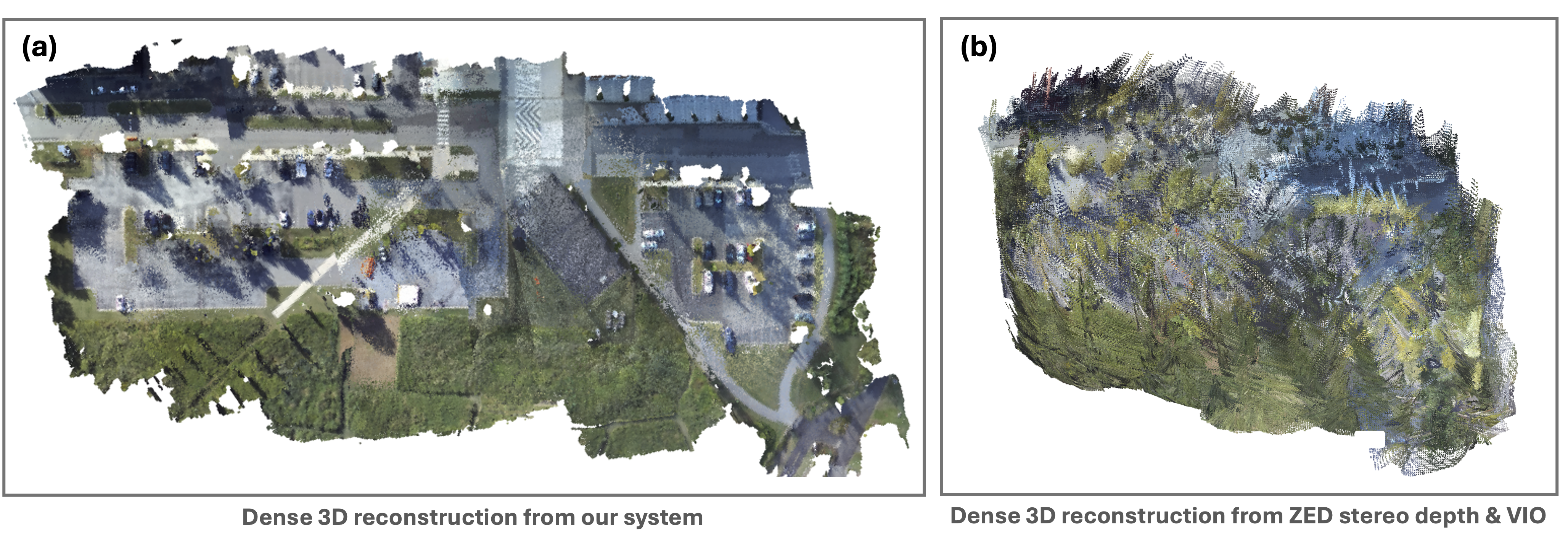}
  \caption{\textbf{Qualitative comparison between dense 3D reconstruction from HALO (a) and from the ZED camera (b)}.
  }
  \label{fig:zed}
  \vspace{-0.3cm}
\end{figure}

\section{Real-world Experiments}

We next validated HALO on a custom-built quadrotor platform. It is equipped with an NVIDIA Jetson AGX Orin onboard computer, a ZED 2i camera, a VectorNav VN-100T IMU, a u-blox ZED-F9P GNSS, and an ARKV6X flight controller running PX4 1.16.
We implement our framework in ROS 2 Humble.

\medskip \noindent \textbf{Setup}
From safety considerations, we operated the robot within a pre-defined region and used a waypoint graph from a satellite image at a desired spatial density. Waypoints within the operating region that are infeasible are rejected.
We use Air-router~\cite{cladera2024enabling} to interface with the flight controller for waypoint tracking in GPS coordinates, and remap the planned path to the nearest reachable waypoint.
Fig.~\ref{fig:experiment} (a) illustrates the set of feasible GPS waypoints overlaid on the satellite image. 

To obtain a stable initial scale estimate for mapping, each experiment is initialized by commanding the quadrotor to perform a fixed-distance motion after takeoff.
In all the experiments, we limit the maximum speed of the quadrotor to 2 m/s. We used the state estimator from~\cite{cladera2025air} to fuse GPS observations with those from the IMU. The desired altitude of the robot is set to 40 m off the ground.

We carried out real-world experiments in two different areas (Fig.~\ref{fig:experiment}) with tasks of varying difficulties (Tab.~\ref{tab:real-world}).
We ran several missions in each area, where each mission can have multiple tasks.

\medskip \noindent \textbf{Results and Analysis}
Fig.~\ref{fig:experiment} (b) shows the geometric map built onboard the robot with the robot trajectory in red. As illustrated, even at an altitude of 40 m, our system consistently captures the 3D geometry of objects and buildings to metric scale.
Fig.~\ref{fig:zed} compares the maps generated by HALO with those from stereo depth and state estimation using the ZED SDK (we used the long-range version with the 4 mm lens).
This highlights the limitations of existing off-the-shelf solutions for real-time high-altitude mapping.

Tab.~\ref{tab:real-world} presents quantitative results from real-world experiments.
Since the commanded waypoints are restricted to the sampled waypoint set, the ground truth path for mission completion is the shortest path computed on the waypoint graph. HALO consistently achieves a competitive ratio of at least 0.35 across tasks of varying difficulty and exceeds 0.8 when the robot has previously encountered relevant information about the task.
This demonstrates the effectiveness of HALO over significant distances and across different tasks in real-world settings.

\begin{table}
\centering
\caption{\textbf{Quantitative results of real-world experiments. Experiments are labeled by Area (A), Experiment (E), and Task (T).}}
\resizebox{\columnwidth}{!}{%
\begin{tabular}{cc c rrr}
    \toprule
    \multicolumn{2}{c}{\multirow{2}{*}{\textbf{Experiment}}}
    & \multirow{2}{*}{\textbf{Task}} 
    & \multirow{2}{*}{\textbf{Distance [m]}}
    & \multirow{2}{*}{\textbf{GT [m]}}
    & {\textbf{Competitive}} \\
    \multicolumn{2}{c}{} & & & & {\textbf{Ratio $\uparrow$}} \\
    \midrule
    A1-E1 & T1 & find a pedestrian crossing & 289.38 & 100.70 & 0.35  \\ \hline
    \multirow{3}{*}{A1-E2} & T1 & find a pedestrian crossing & 240.67 & 100.70 & 0.42 \\ 
     & T2 & find the pool & 139.86 & 114.83 & 0.82 \\ 
     & Total & – & 380.53 & 215.53 & 0.57 \\
     \midrule
    \multirow{3}{*}{A1-E3} & T1 & find a car in the parking lot & 133.40 & 78.56 & 0.59 \\ 
     & T2 & find a roadblock & 45.19 & 35.72 & 0.79 \\ 
     & Total & – & 178.59 & 114.28 & 0.64 \\
     \midrule
    A2-E1 & T1 & find a car in the parking lot & 98.16 & 53.32 & 0.54 \\ \hline
    \multirow{3}{*}{A2-E2} & T1 & find the construction site & 145.26 & 71.40 & 0.49 \\ 
     & T2 & find the red car & 72.88 & 62.96 & 0.86 \\ 
     & Total & – & 218.14 & 134.36 & 0.62 \\
     \bottomrule
\end{tabular}
}
\label{tab:real-world}
\vspace{-0.3cm}
\end{table}

\section{Discussion}

\uline{Flexibility with Task Specification}
Through both simulation and real-world experiments, we show that HALO is able to efficiently complete semantic exploration tasks.
As the metric-semantic map is built incrementally, the planner leverages semantic task relevancy to explore areas that are likely to be useful for completing the task.
HALO achieves a high competitive ratio to the ground truth trajectory across a wide range of tasks.

We also show that our system is able to use the dense language-embedded semantic map to adapt to a range of subsequent tasks.
For instance, in experiment A1-E3 in Tab.~\ref{tab:real-world}, we considered the scenario where the robot is tasked with searching for a vehicle in the parking lot.
Upon successful completion of this first task, the operator notices that there were some barriers set up around the parking lot and would like to check if the exit is blocked.
The robot is then given a second task to inspect the roadblock.
Unlike other approaches that maintain sparse object-level maps with closed-set semantics, our dense semantic feature map enables flexibility with task specifications to accommodate new tasks on the fly, even when the semantics of interest are not known prior to starting the mission or as the map is being constructed online.

\uline{Task Termination}
We used a Qwen2.5-3B~\cite{yang2025qwen3} vision-language model (VLM) to determine task completion by querying the input images with the task description.
However, our framework is agnostic to the choice of termination condition and can be integrated with alternative mechanisms such as an object detector, a higher-level planner or a human operator.
For instance, in tasks that require geometric coverage, the system can terminate once no new frontiers are detected.
When the task objective is ambiguous and multiple valid goals exist, the user can determine when to terminate the current task.

\uline{Computational Requirements}
HALO requires an average of 12 GB of unified memory. We currently embed dense language features into a 2D grid map.
Maintaining higher resolution gridmaps or introducing an additional dimension limits its scalability to large environments.
This can be addressed by introducing submaps and offloading inactive maps to disk to improve memory efficiency.

VGGT demonstrates strong potential for dense 3D reconstruction using aerial imagery.
Although it introduces some latency, we have shown that these systems can support real-time mapping after appropriate modifications.
High-resolution images can enable high-fidelity 3D reconstruction but will add to computational challenges.
It is an interesting direction for future work to balance the fidelity of feed-forward 3D reconstruction approaches with their inherent computational challenges for onboard and real-time execution on autonomous platforms.

\bibliography{literature}

\end{document}